\documentclass[11pt]{article}

\usepackage[preprint]{acl}

\usepackage{times}
\usepackage{latexsym}

\usepackage[T1]{fontenc}
\usepackage[utf8]{inputenc}

\usepackage{microtype}

\usepackage{inconsolata}

\usepackage{graphicx}

\usepackage{amsmath,amssymb,amsfonts}

\usepackage{multirow}
\usepackage{booktabs}
\usepackage[table]{xcolor} % Crucial for \cellcolor to work

\definecolor{emoAngry}{HTML}{C44536}
\definecolor{emoHappy}{HTML}{2D936C}
\definecolor{emoNeutral}{HTML}{4B5563}
\definecolor{emoSad}{HTML}{3A6EA5}

\newcommand{\cmark}{\checkmark}
\newcommand{\xmark}{\textcolor{red}{$\times$}}

\usepackage{xcolor}
\usepackage{colortbl}
\usepackage{tcolorbox}
\definecolor{speech}{HTML}{2E68DB}
\definecolor{transcript}{HTML}{7F00FF}
\definecolor{context}{HTML}{009000}
\definecolor{prompt}{HTML}{666666}
\definecolor{label}{HTML}{FFB570}
\definecolor{labelset}{HTML}{B8860B}

\newenvironment{promptbox}{
  \begin{tcolorbox}[
    colback=black!1,
    colframe=black!25,
    boxrule=0.4pt,
    arc=2mm,
    left=2mm,right=2mm,top=1mm,bottom=1mm,
  ]
  \ttfamily\footnotesize
  \setlength{\tabcolsep}{0pt}
  \renewcommand{\arraystretch}{1.15}
}{
  \end{tcolorbox}
}

\title{From Generative to Discriminative: Lightweight Adaptation of SpeechLLMs for Multimodal Emotion Recognition}

\author{
  \textbf{Hasindri Watawana\textsuperscript{1,2}},
  \textbf{Sergio Burdisso\textsuperscript{1}},
  \textbf{Esaú Villatoro-Tello\textsuperscript{1}},
  \textbf{Manjunath K E\textsuperscript{3}},
\\
  \textbf{Kadri Hacioglu\textsuperscript{3}},
  \textbf{Petr Motlicek\textsuperscript{1,4}},
  \textbf{Andreas Stolcke\textsuperscript{3}}
\\
\\
  \textsuperscript{1}Idiap Research Institute, Switzerland,
  \textsuperscript{2}EPFL, Switzerland,
\\
  \textsuperscript{3}Uniphore, USA \& India,
  \textsuperscript{4}Brno University of Technology, Czech Republic
\\
  \small{
    \textbf{Correspondence:} \href{mailto:hasindri.watawana@idiap.ch}{hasindri.watawana@idiap.ch}
  }
}

\begin{document}
\maketitle
\begin{abstract}
SpeechLLMs can recognise emotion by generating a label token, but this approach relies on the LLM's existing output representations. We compare this generative approach with a discriminative alternative that uses a linear classification head on top of the LLM's hidden states. Because most multimodal emotion recognition studies evaluate with ground-truth transcripts, whereas deployed systems receive ASR output, we run every comparison with both, using a compact speechLLM (frozen Whisper-medium encoder, 3B LLM) on IEMOCAP and MELD. The classification head improves Macro F1 in 10 of 14 configurations. On IEMOCAP, it improves every configuration using ASR transcripts, reaching 76.97\% Macro F1, with a significant gain when using speech alone. On MELD, where ASR transcripts have a 32.38\% word error rate, it helps with speech alone but not with ASR transcripts. The benefit thus depends on both the dataset and the transcript source: in 4 of the 6 configurations that use text, the head's effect changes sign between ground-truth and ASR transcripts. Conclusions drawn with ground-truth transcripts therefore need not hold with ASR transcripts. Finally, comparing the learned class vectors with the LLM's vocabulary shows that, on IEMOCAP and MELD, their nearest tokens are mostly emotion-related words rather than explicit emotion labels.
% SpeechLLMs have shown strong potential for emotion recognition, yet they read the predicted emotion off a generative decoder not suited for classification: it can emit labels outside the target set and favors frequent classes. We propose a discriminative adaptation that reads the final prompt token's hidden state through a classification head, producing a label in one forward pass without modifying the backbone. Because this readout starts from the hidden state the model would otherwise decode, it gives a controlled comparison of generative and discriminative inference in an otherwise identical speechLLM. We keep the head a single linear layer, trading little accuracy for interpretability: each emotion becomes one direction in the LLM output token space, revealing associated tokens. On IEMOCAP, across two speechLLM architectures, it improves Macro F1 and removes hallucinations, with largest gains on realistic ASR transcripts. Our analysis reveals that these emotion directions encode indirect associations mirroring biases in web-scale text.
\end{abstract}

% Keywords: Speech Emotion Recognition, SpeechLLM, Discriminative Adaptation

\section{Introduction}
\label{section:Intro}

Speech Emotion Recognition (SER) supports applications in human--computer interaction, conversational analytics, dialogue systems and mental health monitoring \citep{gross1995emotion,gross2019mental,elsayed2022speech,feng2023end,singla2024emotion}. Despite substantial progress \citep{al2023speech,wang25k_interspeech,chaves2026deep}, it remains challenging because emotion is conveyed through complementary acoustic, visual and linguistic cues, which has motivated Multimodal Emotion Recognition (MER) \citep{10109845,10888112,10889998}. Speech Large Language Models (SpeechLLMs) are a promising basis for MER, as they combine the linguistic knowledge of a pretrained LLM with acoustic evidence. Recent systems add paralinguistic encoders \citep{morais25_interspeech}, cross-attention fusion \citep{du2025eaa}, chain-of-thought distillation \citep{mai25c_interspeech} or emotion-specific encoder pipelines \citep{x_lance_slam_llm_2024}, at the cost of extra modules, losses or training stages.

In this work, we study how emotion labels are predicted from the model's hidden representations. At the answer position, generating a label can be viewed as linear classification, where each class is represented by the embedding of its label word. Instead, we use a linear classification head that takes the same hidden state but learns its own class vectors. Although such heads are commonly used with LLMs \citep{li2023label,wen2026beyond}, our contribution is not the classification head itself but an investigation of when it improves speech emotion recognition and which input representations benefit from it.

Most MER studies use reference transcripts, whereas a deployed system receives ASR output with errors \citep{li2024speech,kyung24_interspeech}, we treat ASR transcripts as the main condition and repeat every configuration with reference transcripts to test whether conclusions carry over. 

We use a compact speechLLM (a frozen Whisper-medium encoder, a two-layer projector and a 3B LLM with LoRA), so that differences come from the readout rather than from added modules. We evaluate on IEMOCAP \citep{busso2008iemocap} and MELD \citep{poria-etal-2019-meld} with speech alone, adding its transcript, and adding the preceding utterance as context; with all three inputs, we also freeze the LLM. Because the head is linear, its class vectors can be compared with the LLM's vocabulary \citep{dar-etal-2023-analyzing}.

% We conduct an extensive evaluation on the widely used IEMOCAP emotion recognition benchmark, under multimodal settings using speech and text data.

Our main contributions are as follows:
(1) \textbf{A controlled evaluation of the readout.} The head improves Macro F1 in 10 of the 14 configurations we report. In the IEMOCAP ASR setting it improves all three input configurations, reaching 76.97 Macro F1 (76.80 WA and 77.50 UA, above prior ASR-based systems), with a significant gain from speech alone. On MELD it helps with speech alone, but not with ASR transcripts.
(2) \textbf{Evidence that reference transcripts can mislead.} The sign of the head's effect differs between reference and ASR transcripts in 4 of the 6 configurations that use text, so conclusions drawn from reference transcripts do not carry over. ASR errors also shrink what text adds: on MELD (32.38\% WER), text adds 5.7 Macro F1 over speech alone with ASR transcripts, against 13.9 with reference transcripts.
(3) \textbf{A vocabulary-space analysis of the head.} On IEMOCAP, the nearest tokens of the class vectors recur across folds and are mostly emotion-laden words rather than emotion labels (e.g., \emph{argument} and \emph{interrupt} for anger). On MELD, the vectors for fear, surprise and disgust point to explicit emotion words (e.g., \emph{terrified}, \emph{astonished}, \emph{disgusted}). We report these as associations, not explanations.
%(1) \textbf{Interpretability through linearity:} We show that parameterizing the classifier as a single linear layer turns it into an interpretability handle: because each emotion is one direction in the LLM hidden space, its class vector can be projected through the model's own unembedding matrix to recover the tokens it is associated with. This exposes that the learned emotion directions encode culturally grounded associations that mirror biases present in large-scale web text. %---a property a non-linear head would forfeit (Section~\ref{subsection:interpretability}).(2) \textbf{A controlled comparison of generative vs.\ discriminative readout:} Because the generated label and the discriminative prediction are read from the \emph{same} final-token hidden state, we obtain a controlled comparison of the two readout mechanisms on an otherwise identical, frozen speechLLM. Discriminative supervision removes the risk of hallucinated (invalid) label tokens and rebalances predictions toward minority emotions---improving Macro\,F1---where generative decoding favors the majority class.(3) \textbf{Robustness under realistic transcripts:} Evaluating with automatically generated (ASR) transcripts rather than the oracle transcripts assumed by most MER studies, we find the advantage of discriminative readout is largest when input quality is degraded, indicating concrete benefit for deployment.(4) \textbf{Cross-architecture validation:} We confirm these effects across two speechLLM pipelines (single- and dual-encoder), showing they are not specific to a single architecture.

\begin{figure*}
    \centering
    \includegraphics[width=0.65\linewidth]{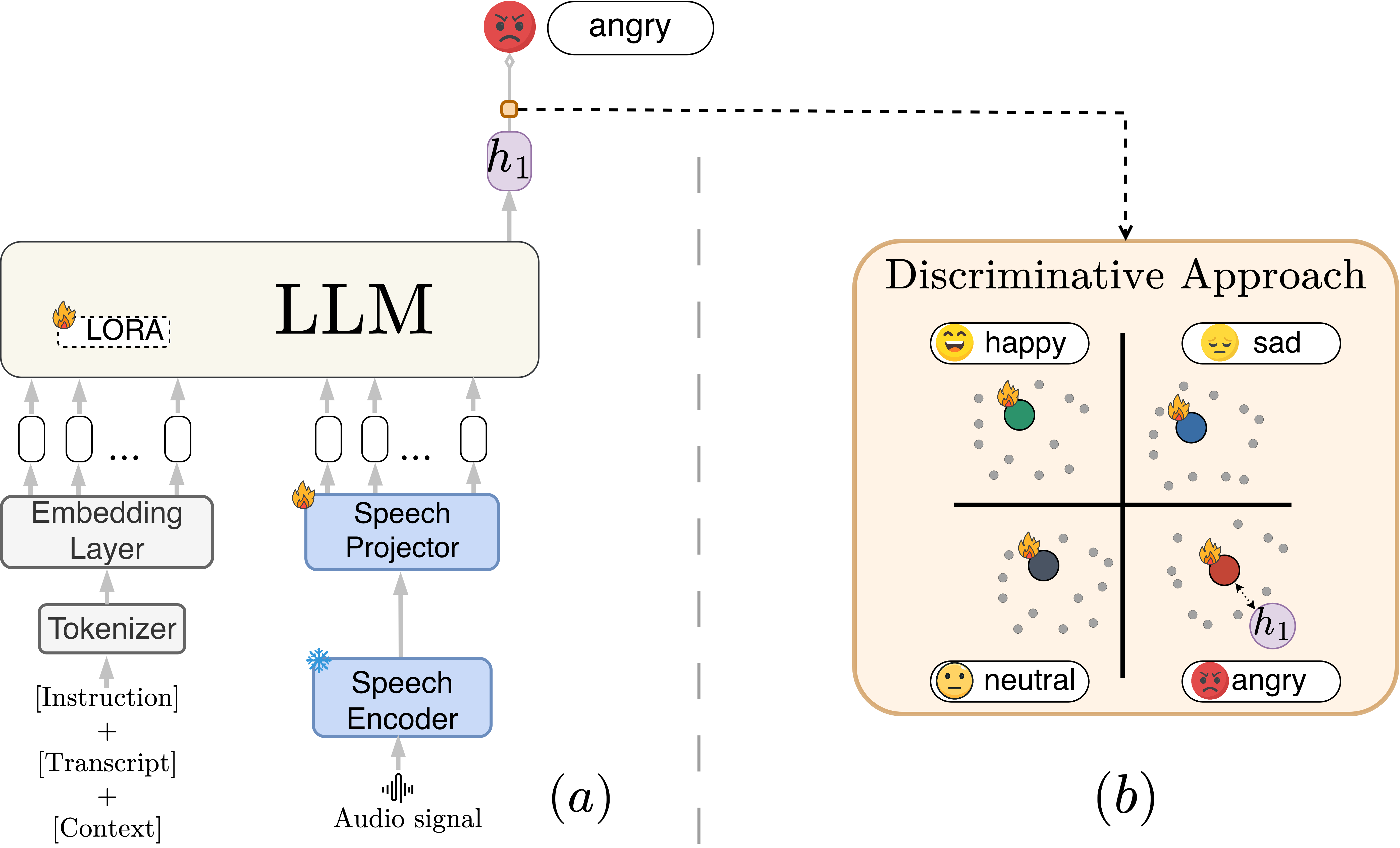}
    \caption{\textbf{General view of the proposed framework.} (a) \textbf{Architecture:} a frozen speech encoder, a trainable speech projector, and a LoRA-adapted LLM generate a single hidden state, $h_1$ (hidden state of the final prompt token). (b) \textbf{Discriminative head:} emotion classification is performed directly from $h_1$ by learning one vector per emotion (green, blue, gray, and red circles with flame icons) in the token embedding space. These vectors act as learnable centroids that represent emotion-specific clusters of token hidden states. At inference time, the predicted emotion corresponds to the centroid with the highest dot-product similarity to $h_1$. In this illustrated example, $h_1$ is closest to the red centroid, resulting in the prediction \texttt{angry}.}
    \label{fig:general_pipeline}
\end{figure*}

\section{Related Work}

Large spoken language models vary widely in how the speech modality is integrated with the LLM: how speech representations are learned and aligned with a (text-based) LLM. Our method follows the connector paradigm, in which representations from a pretrained speech encoder are bridged to the LLM through a lightweight adapter trained for the target task, while the backbone remains frozen. This preserves the LLM's pretrained knowledge and instruction-following ability, confining adaptation to a small set of trainable parameters.

Closer to our setting is a line of work that adapts a \emph{frozen} LLM as a speech emotion classifier through a lightweight connector. Bellver-Soler et al.~\cite{bellversoler2026ser} condense speech to a single acoustic token via attentive pooling, project it into a frozen LLM alongside a text instruction, and obtain the prediction by applying a softmax over the next-token logits of the emotion-label tokens; they report that the frozen LLM outperforms an MLP head on several corpora including IEMOCAP. Both this approach and the discrete-token method of Calbucura et al.~\cite{calbucura2026simple} thus derive the label from the model's own (constrained) generative machinery, leaving the backbone essentially unchanged. In contrast, we append a discriminative linear head to the final hidden state, trained with a classification objective, and directly contrast this readout against generative label decoding on the \emph{same} frozen speechLLM. Both restricting the label-token logits~\cite{bellversoler2026ser} and our head eliminate hallucinated labels, but by different means; we further adapt the backbone with LoRA and condition on speech, transcript, and dialogue context, rather than a single transcript-free acoustic token.

A related and still-open question is what these adapted representations encode. Bellver-Soler et al.~\cite{bellversoler2026ser} observe that the projected acoustic token does not lie near emotion words yet still supports accurate classification, and leave a characterization of this embedding geometry to future work; analyses of spoken LLMs similarly note that connectors make utterance-level semantics linearly accessible without a one-to-one lexical correspondence~\cite{11434765}. We address this from the output side: as our head is linear, projecting each class vector through LLM unembedding recovers the associated tokens per emotion, revealing that the learned emotion directions align not with literal affect words but with indirect associations. Where prior work asks whether the \emph{input} token sits near emotion words, we expose what the \emph{class directions} themselves point to.

\section{Methodology}
\label{section:methodology}

\subsection{SpeechLLM Framework}
\label{section:speechLLM}

We employ a speechLLM for emotion understanding as it combines broad world knowledge and reasoning ability from pretrained LLMs with speech--text alignment, which grounds these linguistic priors in acoustic evidence. Our architecture is inspired by the SLAM-ASR framework \cite{x_lance_slam_llm_2024,ma2024embarrassingly}, originally proposed for automatic speech recognition (ASR). This architecture connects a frozen pre-trained speech encoder to a frozen pre-trained LLM via a lightweight projector network (Refer Figure~\ref{fig:general_pipeline}). Given a speech input, the encoder produces a sequence of frame embeddings $\mathbf{X} = (\mathbf{x}_1, \dots, \mathbf{x}_T) \in \mathbb{R}^{T \times d_s}$, where $T$ is the sequence length and $d_s$ the encoder hidden dimension. To reduce the sequence length, $k$ consecutive frames are concatenated, yielding a downsampled sequence $\mathbf{Z} \in \mathbb{R}^{\lfloor T/k \rfloor \times kd_s}$. This sequence is then passed through a two-layer MLP projector that maps the speech representations into the LLM input space,
\begin{equation}
\mathbf{E} = \text{ReLU}\left( \mathbf{Z}\, \mathbf{W}_{1} \right) \mathbf{W}_{2}
\label{eq:speech_projector}
\end{equation}
where $\mathbf{W}_1 \in \mathbb{R}^{kd_s \times d_h}$ and $\mathbf{W}_2 \in \mathbb{R}^{d_h \times d_l}$, %with $d_h$ the projector hidden dimension and $d_l$ the LLM input dimension.
with $d_h$ the projector hidden dimension and $d_l$ the LLM model dimension (shared by the input embedding, hidden state, and unembedding in the decoder-only LLMs we use).
The resulting sequence $\mathbf{E} \in \mathbb{R}^{\lfloor T/k \rfloor \times d_l}$ serves as the speech prompt to the LLM, replacing the \textcolor{speech}{\texttt{\textbf{\{speech\}}}} token in the input prompt prior to the forward pass. To adapt this architecture for MER, we replace the original ASR prompt with a multimodal instruction template introducing two additional placeholders, \textcolor{transcript}{\texttt{\textbf{\{transcript\}}}} and \textcolor{context}{\texttt{\textbf{\{context\}}}}, alongside \textcolor{speech}{\texttt{\textbf{\{speech\}}}}:

\begin{promptbox}
\ttfamily\footnotesize
\textbf{\textcolor{prompt}{<s>USER: You are a classifier. Predict the emotion of the CURRENT utterance using dialogue context, transcript, and speech. Valid labels (output EXACTLY one): \textcolor{labelset}{\{label\_list\}}. Output format: (LABEL). Do NOT output any other words, punctuation, explanations, or multiple labels.}}\\
\textbf{\textcolor{prompt}{Context (previous utterance):} \textcolor{context}{\{context\}}\textcolor{prompt}{.}}
\textbf{\textcolor{prompt}{Current transcript:} \textcolor{transcript}{\{transcript\}}\textcolor{prompt}{.}}
\textbf{\textcolor{prompt}{Speech:} \textcolor{speech}{\{speech\}}\textcolor{prompt}{.}}
\textbf{\textcolor{prompt}{Answer:}}\\
\textbf{\textcolor{prompt}{ASSISTANT:} \textit{\textcolor{label}{\{label\}}}\textcolor{prompt}{</s>}}
\end{promptbox}
%Unlike the original SLAM-ASR work, which uses a single \textcolor{speech}{\texttt{\textbf{\{speech\}}}} token since the task is ASR, we adopt a richer prompt template suited for emotion classification:

% \noindent
% \fbox{%
% \scriptsize
%   \parbox{\dimexpr\columnwidth-2\fboxsep-2\fboxrule\relax}{%
%     \ttfamily
%     You are a classifier. Predict the emotion of the CURRENT utterance using dialogue context, transcript, and speech. Valid labels (output EXACTLY one): Angry, Happy, Sad, Neutral. Output format: (LABEL). Do NOT output any other words, punctuation, explanations, or multiple labels. Context (previous utterance): \textbf{<context>}. Current transcript: \textbf{<text>}. Speech: \textbf{<speech>}. Answer:
%   }%
% }\\

This allows the LLM to perform multimodal classification over both acoustic (\textcolor{speech}{\texttt{\textbf{\{speech\}}}}) and linguistic (\textcolor{transcript}{\texttt{\textbf{\{transcript\}}}}) inputs, conditioned on dialogue history (\textcolor{context}{\texttt{\textbf{\{context\}}}}). \textcolor{labelset}{\texttt{\textbf{\{label\_list\}}}} is the set of valid emotion labels for the target dataset  and (\textcolor{label}{\texttt{\textbf{\{label\}}}}) is the target emotion label of a given utterance. 
% While including both speech and its transcript as input may seem redundant, each modality captures complementary information, and as shown in Tables ~\ref{tab:main_results_iemocap} \&~\ref{tab:main_results_meld}, combining them consistently improves performance. We conduct experiments with and without each placeholder to assess their individual contributions to emotion classification. 
In this work, \textit{text} refers to the transcription of the input speech utterance, while \textit{context} consists of a single preceding utterance.

In the \textit{generative} formulation, the predicted emotion label is obtained by reading the tokens generated by the LLM conditioned on the full prompt,
\begin{equation}
\hat{c} = \text{LLM}\left(\mathbf{E},\, \mathbf{P}\right)
\label{eq:llm_classification}
\end{equation}
where $\mathbf{P}$ is the textual prompt encoding the \textit{text} and \textit{context} fields, $\mathbf{E}$ are the speech prompt embeddings, and $\hat{c}$ the generated emotion label.

\subsection{From Generative to Discriminative}
\label{section:CLS_head}

In the \textit{generative} formulation, model reads the predicted emotion off the LLM by decoding: conditioned on the full prompt, the model decodes tokens autoregressively until an end-of-sequence token, and the output string is parsed into one of the labels (Equation~\ref{eq:llm_classification}). Concretely, at decoding step $i$, the hidden state $\mathbf{h}_i \in \mathbb{R}^{d_l}$ % is projected through the unembedding matrix $\mathbf{W}_{lm} \in \mathbb{R}^{d_l \times |\mathcal{V}|}$ and the highest-scoring token $t_i$ is emitted,
is projected through the unembedding matrix $\mathbf{W}_{lm} \in \mathbb{R}^{d_l \times |\mathcal{V}|}$, whose $v$-th column $\mathbf{w}^{lm}_{v} \in \mathbb{R}^{d_l}$ is the output embedding of token $v$, and the highest-scoring token $t_i$ is emitted,
\begin{equation}
t_i = \underset{v \in \mathcal{V}}{\arg\max}\ \mathbf{h}_i\, \mathbf{w}^{lm}_{v}.
\label{eq:next_token}
\end{equation}
This generative readout is a poor fit for classification: it can emit tokens outside the label set (\textit{e.g.,} hallucinations) and, deciding in a space optimized for next-token prediction, may inherit the language-modeling objective's bias toward frequent labels.

A standard remedy is to treat the decoder as an encoder: pool the hidden state of the last input token, which under causal masking has attended to the entire sequence, and attach a classification head~\cite{li2023label}. Prior adaptations of this kind typically use a higher-capacity, non-linear head~\cite{wen2026beyond}, which favors accuracy but spreads each class across many entangled parameters, leaving what the head learns about an emotion opaque. We instead constrain the head to a \emph{single linear layer}, trading a little capacity for interpretability: a linear head represents each emotion by one vector in the LLM hidden space that can later be examined directly. Concretely, we read the final prompt token's last hidden state $\mathbf{h}_1$ (the position from which the first answer token would be generated) through this head over the emotion classes $\mathcal{C}$ rather than through the unembedding (Figure~\ref{fig:general_pipeline}). The classification head (CLS head) is defined as follows,
\begin{equation}
\hat{c} = \underset{c \in \mathcal{C}}{\arg\max}\ \mathbf{h}_1\, \mathbf{w}^{\text{CLS}}_{c},
\label{eq:eq4}
\end{equation}
with $\mathbf{W}_{\text{CLS}} \in \mathbb{R}^{d_l \times |\mathcal{C}|}$, whose $c$-th column $\mathbf{w}^{\text{CLS}}_{c} \in \mathbb{R}^{d_l}$ is the direction associated with emotion $c$ (mirroring $\mathbf{w}^{lm}_{v}$ above). The decision requires a single forward pass, eliminates invalid labels by construction, and rebalances predictions toward minority emotions (Section~\ref{section:results}).
% WHAT ABOUT UNEMBEDDING FUNCTION AND TRAINING LOSS
The head is trained exactly as the token layer: a softmax over the class scores followed by cross-entropy against the ground-truth emotion, the same objective the LLM uses over its vocabulary, only with the labels drawn from $\mathcal{C}$ instead of $\mathcal{V}$.

This parallelism is deliberate. Keeping the head linear casts Equation~\ref{eq:eq4} in the same form as the unembedding step of Equation~\ref{eq:next_token}, a matrix applied to $\mathbf{h}$ followed by an $\arg\max$, so the two differ only in their output space: one column per token in $\mathbf{W}_{lm}$, one per emotion in $\mathbf{W}_{\text{CLS}}$. Because both act on the same hidden state, their columns are comparable: each is trained to align with the hidden states that select it, placing emotion vectors and token embeddings in one shared space. Discriminative training then aligns each emotion column with the hidden states diagnostic of its emotion and, through them, with the token embeddings those states would decode, so the column settles as a learned \emph{centroid} among the associated tokens (Figure~\ref{fig:tsne}). Comparing an emotion column against the token embeddings in $\mathbf{W}_{lm}$ recovers the most similar tokens, exposing the associations the model has tied to each emotion (Section~\ref{sec:interpretability}).

\section{Experimental Setup}
\label{section:experiments}

\subsection{Datasets}

% We use the IEMOCAP database \cite{busso2008iemocap} for all experiments, a widely adopted multimodal benchmark comprising audio, transcripts, video, and motion capture recordings of dyadic interactions across five sessions, each involving a distinct pair of actors. We follow the standard Leave-One-Session-Out cross-validation protocol for speaker-independent evaluation. Although IEMOCAP contains ten emotion categories, we adopt the common 4-class formulation \cite{ma2024emobox}, merging Excited into Happy, yielding Angry, Happy, Sad, and Neutral across 5,531 utterances (1,103 / 1,636 / 1,084 / 1,708 respectively).

We use the IEMOCAP \cite{busso2008iemocap} and MELD \cite{poria-etal-2019-meld} datasets. IEMOCAP is a widely adopted multimodal benchmark comprising audios, transcripts and videos of dyadic interactions across five sessions, each involving a distinct pair of actors. We follow the standard Leave-One-Session-Out (LOSO) 5-fold cross-validation protocol. Although IEMOCAP contains ten emotion categories, we adopt the common 4-class formulation \cite{ma2024emobox}, merging Excited into Happy, yielding Angry, Happy, Sad, and Neutral across 5,531 utterances (1,103 / 1,636 / 1,084 / 1,708 respectively). MELD comprises 13,708 multi-party utterances from the TV series \textit{Friends}, annotated with seven emotions (Neutral, Surprise, Fear, Sadness, Joy, Disgust, Anger). We use the original 7-class formulation and the official train/dev/test split (9,989 / 1,109 / 2,610), reporting test-set scores. 

\subsection{Unimodal Baselines}
\label{section:SER_baselines}

\noindent$\bullet$ \emph{\textbf{Speech Baselines with Speech Foundation Models.}} Speech foundation models (SFMs) are large-scale neural networks pretrained on diverse audio data to learn general-purpose representations of speech. These representations are commonly used for downstream audio tasks. Adapting SFMs to SER requires task-specific fine-tuning. For speech baselines, we replicate the MLP probing setup of the EmoBox SER benchmark~\cite{ma2024emobox}\footnote{\url{https://github.com/emo-box/EmoBox}}.

\noindent$\bullet$ \emph{\textbf{Text Baselines with Zero-Shot LLM.}} We evaluate the emotion understanding ability of state-of-the-art LLMs from ground truth transcripts. 

\noindent$\bullet$ \emph{\textbf{Speech Baselines with Multimodal LLMs.}} Multimodal LLMs (MLLMs) are models that reason across multiple data modalities, integrating audio and visual understanding into a text-pretrained LLM backbone. We evaluate several state-of-the-art MLLMs on emotion recognition using speech input only, out of the box, without any fine-tuning.

%IEMOCAP MAIN RESULTS TABLE
\begin{table*}[t!]
% continuous vertical rules: remove booktabs' gaps above/below horizontal rules,
% and restore the lost breathing room with extra row height
\setlength{\aboverulesep}{0pt}
\setlength{\belowrulesep}{0pt}
\setlength{\extrarowheight}{.4ex}
\setlength{\tabcolsep}{6pt}
\renewcommand{\arraystretch}{1.0}
\centering
\footnotesize
\caption{Main results on IEMOCAP. The first two blocks report prior work from the literature; the third reports our speechLLM. \textbf{+ CLS} denotes our speechLLM with the classification head appended. All methods take \textit{audio} as input. \textbf{ASR} characterises the textual input: \cmark{} for ASR transcripts, \xmark{} for ground-truth transcripts, and ``--'' when the method uses no text data at all. \textbf{Context} indicates whether preceding textual context is provided (in our speechLLM, the context is ASR-transcribed as well). Our results are mean\,$\pm$\,std over the 5 folds; \textbf{bold} marks the best result in each input configuration. ``$*$'' and ``$\dagger$'' represent statistically significant (Paired Single-tailed $t$-test, $p=0.05$) improvements over Sys.1 and Sys.2, respectively.}
\label{tab:main_results_iemocap}
\begin{tabular}{l|l|c|c|c|c|c}
\toprule
\multirow{2}{*}{\textbf{ID}} & \multirow{2}{*}{\textbf{Method}}
& \multicolumn{2}{c|}{\textbf{Textual Inputs}}
& \multicolumn{3}{c}{\textbf{IEMOCAP}} \\
\cmidrule(lr){3-4} \cmidrule(lr){5-7}
& & \textbf{ASR} & \textbf{Context}
& \textbf{WA} & \textbf{UA} & \textbf{Macro F1} \\
\midrule

% ---- Block 1: prior work ----
\rowcolor{gray!15}
& \multicolumn{6}{l}{\textit{Prior work}} \\
\midrule
& emotion2vec \cite{maemotion2vec} & -- & -- & 71.79 & -- & -- \\
& ENT \cite{shen2024emotion}       & -- & -- & 72.43 & 73.88 & -- \\
& Li et al. \cite{li2024speech}    & \cmark & -- & 74.66 & -- & -- \\
\midrule

% ---- Block 2: speechLLM-based ----
\rowcolor{gray!15}
& \multicolumn{6}{l}{\textit{SpeechLLM-based}} \\
\midrule
& SALMONN \cite{tang2024salmonn}          & -- & -- & -- & 67.00 & -- \\
& EMOQ \cite{yang2025emoq}                & \xmark & -- & 74.40 & 74.50 & -- \\
& Kyung et al. \cite{kyung24_interspeech} & \cmark & -- & 76.11 & 77.16 & -- \\
& Chen et al. \cite{chen25i_interspeech}  & -- & -- & -- & 77.60 & -- \\
\midrule

% ---- Block 3: ours ----
\rowcolor{gray!15}
& \multicolumn{6}{l}{\textit{Ours (ASR transcripts)}} \\
\midrule
1 & Ours
& -- & --
& 69.86\,{\scriptsize$\pm$2.36} & 70.50\,{\scriptsize$\pm$2.43} & 70.06\,{\scriptsize$\pm$2.63} \\
2 & Ours (+ CLS)
& -- & --
& \textbf{72.35}\,{\scriptsize$\pm$1.80} & \textbf{73.14}\,{\scriptsize$\pm$2.14} & \textbf{72.80}$^{*}$\,{\scriptsize$\pm$2.06} \\
\cmidrule{1-7}
3 & Ours
& \cmark & --
& 74.52\,{\scriptsize$\pm$3.03} & 75.45\,{\scriptsize$\pm$3.55} & 74.90\,{\scriptsize$\pm$3.27} \\
4 & Ours (+ CLS)
& \cmark & --
& \textbf{75.47}\,{\scriptsize$\pm$2.74} & \textbf{76.26}\,{\scriptsize$\pm$3.13} & \textbf{75.81}$^{\dagger}$\,{\scriptsize$\pm$2.64} \\
\cmidrule{1-7}
5 & Ours
& \cmark & \cmark
& 74.49\,{\scriptsize$\pm$2.99} & 74.34\,{\scriptsize$\pm$5.58} & 74.47\,{\scriptsize$\pm$4.62} \\
6 & Ours (+ CLS)
& \cmark & \cmark
& \textbf{76.80}\,{\scriptsize$\pm$3.66} & \textbf{77.50}\,{\scriptsize$\pm$3.47} & \textbf{76.97}$^{\dagger}$\,{\scriptsize$\pm$3.41} \\
\bottomrule
\end{tabular}
\end{table*}

\subsection{SpeechLLM Implementation}
\label{sec:speechllm_implementation}

% IEMOCAP audio recordings are mono-channel and sampled at 16 kHz, which we use directly as input. Our pipeline employs Whisper-medium encoder\footnote{\url{https://huggingface.co/openai/whisper-medium}} as the speech encoder, and LLaMA-3.2-3B-Instruct as the LLM \cite{dubey2024llama}. To enable parameter-efficient adaptation of the LLM, we adopt LoRA, updating only a limited subset of parameters. We set the LoRA rank to 16, the LoRA scaling factor ($\alpha$) to 32, and apply a dropout rate of 0.05. The base LLM and speech encoder are frozen during training, while the speech projector, CLS head, and LoRA parameters are optimized simultaneously. Model is trained for 10 epochs with a batch size of 6 using AdamW optimizer with an initial learning rate of $1 \times 10^{-4}$, zero weight decay, and 1000 warm-up steps. Training is conducted using mixed precision (bf16) on a single NVIDIA H100 GPU for each experiment.

IEMOCAP and MELD audios are mono-channel and sampled at 16 kHz, which we use directly as input. Our pipeline employs Whisper-medium\footnote{\url{https://huggingface.co/openai/whisper-medium}} as the speech encoder, and LLaMA-3.2-3B-Instruct as the LLM \cite{dubey2024llama}. To enable parameter-efficient adaptation of the LLM, we adopt LoRA, updating only a limited subset of parameters. We set the LoRA rank to 16, scaling factor ($\alpha$) to 32, and apply a dropout rate of 0.05. The base LLM and speech encoder are frozen during training, while the speech projector, CLS head, and LoRA parameters are optimized simultaneously. Models are trained with a batch size of 6 using the AdamW optimizer with an initial learning rate of $1 \times 10^{-4}$, zero weight decay, and 1000 warm-up steps, for 10 epochs on IEMOCAP and 6 epochs on MELD. Training is conducted using mixed precision (bf16) on a single NVIDIA H100 GPU for each experiment.
% For the ASR setting, IEMOCAP transcripts are produced by Whisper-medium fine-tuned separately per fold (10 epochs), so fold $k$ decodes Session $k$, its held-out session, and every utterance is transcribed by a model that never trained on it (WER: 8.06\%). On MELD we use zero-shot Whisper-medium (WER: 32.38\%): a MELD-fine-tuned checkpoint would have been trained on the MELD training split, yielding markedly cleaner transcripts for training than for testing, so zero-shot decoding keeps transcript quality matched across splits.
For the ASR setting, IEMOCAP transcripts are produced by Whisper-medium (WER: 8.06\%). On MELD we use zero-shot Whisper-medium (WER: 32.38\%).

\subsection{Evaluation Metrics}

% We report three standard evaluation metrics commonly used in emotion recognition benchmarks. \textbf{Weighted Accuracy (WA)} is the ratio of correct predictions to total samples, and is therefore influenced by majority classes. \textbf{Unweighted Accuracy (UA)} is the macro-averaged recall computed as the mean of per-class recall scores. \textbf{Macro F1} is the unweighted average of per-class F1 scores. Macro F1 is our primary metric of interest, as it weights all emotion categories equally regardless of their frequency and accounts for both precision and recall.

We report four standard evaluation metrics commonly used in emotion recognition benchmarks. \textbf{Weighted Accuracy (WA)} ratio of correct predictions to total samples (influenced by majority classes). \textbf{Unweighted Accuracy (UA)} macro-averaged recall computed as the mean of per-class recall scores. \textbf{Macro F1} unweighted average of per-class F1 scores. \textbf{Weighted F1 (W-F1)} averaged per-class F1 scores weighted by class support; we report it on MELD for comparability with prior work. All results are reported as mean $\pm$ standard deviation: over the 5 folds of the LOSO protocol on IEMOCAP, and over 3 random runs on MELD.

\section{Results}
\label{section:results}

\begin{table*}[t!]
% continuous vertical rules: remove booktabs' gaps above/below horizontal rules,
% and restore the lost breathing room with extra row height
\setlength{\aboverulesep}{0pt}
\setlength{\belowrulesep}{0pt}
\setlength{\extrarowheight}{.4ex}
\setlength{\tabcolsep}{6pt}
\renewcommand{\arraystretch}{1.0}
\centering
\footnotesize
\caption{Main results on MELD. The first two blocks report prior work from the literature; the last two report our speechLLM with ASR and ground-truth (GT) transcripts, respectively. All methods take \textit{audio} as input except those marked ``$\circ$''; methods marked ``$\ddagger$'' additionally use \textit{video}. \textbf{ASR} characterises the textual input: \cmark{} for ASR transcripts, \xmark{} for GT transcripts, and ``--'' when the method uses no text data at all. \textbf{Context} indicates whether preceding textual context is provided; it is transcribed in the same way as the current utterance. Our results are mean\,$\pm$\,std over 3 random seeds; \textbf{bold} marks the best result in each configuration. ``$*$'' and ``$\dagger$'' represent statistically significant (Paired Single-tailed $t$-test, $p=0.05$) improvements over Sys.1 and Sys.2, respectively.}
\label{tab:main_results_meld}
\begin{tabular}{l|l|c|c|c|c|c}
\toprule
\multirow{2}{*}{\textbf{ID}} & \multirow{2}{*}{\textbf{Method}}
& \multicolumn{2}{c|}{\textbf{Textual Inputs}}
& \multicolumn{3}{c}{\textbf{MELD}} \\
\cmidrule(r){3-4} \cmidrule(l){5-7}
& & \textbf{ASR} & \textbf{Context}
& \textbf{WA} & \textbf{W-F1} & \textbf{Macro F1} \\
\midrule

% ---- Block 1: prior work ----
\rowcolor{gray!15}
& \multicolumn{6}{l}{\textit{Prior work}} \\
\midrule
& emotion2vec \cite{maemotion2vec}                        & --     & --     & 51.88 & 48.70 & -- \\
& DGODE$^{\ddagger}$ \cite{shou-etal-2025-dynamic}        & \xmark & \cmark & --    & 67.20 & -- \\
& DCLF$^{\ddagger}$ \cite{xie2025dual}                    & \xmark & \cmark & 68.37 & 67.11 & -- \\
& DialogueMMT$^{\ddagger\circ}$ \cite{he2025dialoguemmt}  & \xmark & \cmark & 71.19 & 70.66 & -- \\
\midrule

% ---- Block 2: speechLLM- and LLM-based ----
\rowcolor{gray!15}
& \multicolumn{6}{l}{\textit{SpeechLLM- and LLM-based}} \\
\midrule
& AA-SLLM \cite{mai2025aa}                  & --     & --     & --    & 55.23 & --    \\
& EMO-RL \cite{li2025emo}                   & --     & --     & 69.56 & --    & 41.87 \\
& EMOQ \cite{yang2025emoq}                  & \xmark & --     & 67.60 & 66.50 & --    \\
& SpeechCueLLM \cite{wu2025beyond}          & \xmark & \cmark & --    & 67.60 & --    \\
& LaERC-S$^{\circ}$ \cite{fu2025laerc}      & \xmark & \cmark & --    & 69.27 & --    \\
\midrule

% ---- Block 3: ours, ASR transcripts ----
\rowcolor{gray!15}
& \multicolumn{6}{l}{\textit{Ours (ASR transcripts)}} \\
\midrule
1 & Ours
& -- & --
& 51.62\,{\scriptsize$\pm$0.84} & 49.95\,{\scriptsize$\pm$0.87} & 32.09\,{\scriptsize$\pm$0.37} \\
2 & Ours (+ CLS)
& -- & --
& \textbf{51.95}\,{\scriptsize$\pm$0.83} & \textbf{50.45}\,{\scriptsize$\pm$0.80} & \textbf{33.32}$^{*}$\,{\scriptsize$\pm$0.54} \\
\cmidrule{1-7}
3 & Ours
& \cmark & --
& 54.48\,{\scriptsize$\pm$0.90} & \textbf{53.45}\,{\scriptsize$\pm$0.96} & \textbf{37.75}\,{\scriptsize$\pm$0.39} \\
4 & Ours (+ CLS)
& \cmark & --
& \textbf{54.81}\,{\scriptsize$\pm$0.73} & 53.21\,{\scriptsize$\pm$0.93} & {37.12}$^{\dagger}$\,{\scriptsize$\pm$2.08} \\
\cmidrule{1-7}
5 & Ours
& \cmark & \cmark
& 54.27\,{\scriptsize$\pm$0.27} & \textbf{53.21}\,{\scriptsize$\pm$0.28} & \textbf{37.76}\,{\scriptsize$\pm$0.27} \\
6 & Ours (+ CLS)
& \cmark & \cmark
& \textbf{54.53}\,{\scriptsize$\pm$1.25} & 53.08\,{\scriptsize$\pm$1.52} & {36.54}$^{\dagger}$\,{\scriptsize$\pm$2.00} \\
\midrule

% ---- Block 4: ours, ground-truth transcripts ----
\rowcolor{gray!15}
& \multicolumn{6}{l}{\textit{Ours (GT transcripts)}} \\
\midrule
7 & Ours
& \xmark & --
& 62.91\,{\scriptsize$\pm$0.71} & 62.29\,{\scriptsize$\pm$0.53} & 45.97\,{\scriptsize$\pm$0.62} \\
8 & Ours (+ CLS)
& \xmark & --
& \textbf{64.94}\,{\scriptsize$\pm$0.76} & \textbf{63.92}\,{\scriptsize$\pm$0.62} & \textbf{48.59}$^{\dagger}$\,{\scriptsize$\pm$1.47} \\
\cmidrule{1-7}
9 & Ours
& \xmark & \cmark
& 63.98\,{\scriptsize$\pm$0.14} & 63.53\,{\scriptsize$\pm$0.15} & 47.04\,{\scriptsize$\pm$0.50} \\
10 & Ours (+ CLS)
& \xmark & \cmark
& \textbf{66.04}\,{\scriptsize$\pm$0.54} & \textbf{64.81}\,{\scriptsize$\pm$0.66} & \textbf{48.77}$^{\dagger}$\,{\scriptsize$\pm$1.31} \\
\bottomrule
\end{tabular}
\end{table*}

We first establish unimodal baselines: speech baselines with SFMs and out-of-the-box MLLMs, and text baselines with zero-shot LLMs; full results are reported in Appendix~\ref{sec:unimodal_baselines} (Table~\ref{tab:unimodal_baselines}). Whisper-large is the strongest SFM, achieving a Macro F1 of 70.45 on IEMOCAP and 27.80 on MELD, against the 73.11 and 32.95 of the current first-place entry of the official EmoBox leaderboard,\footnote{\url{https://emo-box.github.io/leaderboard1.html}} shown for reference. State-of-the-art MLLMs give the strongest speech baselines on both datasets despite no dataset-specific fine-tuning --- Audio Flamingo 3 reaching 76.69 Macro F1 on IEMOCAP and Qwen3-Omni-30B 43.33 on MELD --- indicating how far general-purpose speech understanding has advanced in LLM-backboned models. The text-only baseline, Gemma3-27b, separates the two corpora: on IEMOCAP it reaches 54.16 Macro F1, well below every SFM and the stronger MLLMs, underscoring the role of prosodic and acoustic information in emotion expression; on MELD it reaches 47.32, above every speech baseline, indicating that MELD's emotional signal resides largely in the words rather than in the audio.

% Table~\ref{tab:main_results} summarizes the main results of our proposed speechLLM-based MER framework. For each configuration, we evaluate two variants: with ($\checkmark$) and without ($\times$) the CLS head. Several consistent trends emerge across settings.
% First, multimodal speech--text inputs yield clear gains over unimodal baselines (with the exception of Audio Flamingo 3), confirming that acoustic and linguistic modalities carry complementary emotional information. Second, adding conversational context (+ context) further improves performance across all configurations, with Macro F1 reaching 78.14, suggesting that discourse-level cues are beneficial for emotion inference.
% Most importantly, the CLS head consistently improves or matches performance on UA and Macro F1 (which weight all classes equally) across all input settings and both encoder configurations, while the variant without CLS head tends to favor WA, indicating a bias toward majority classes at the expense of minority emotion recognition.

Table~\ref{tab:main_results_iemocap} reports our main IEMOCAP results. To simulate the real-world setting, all configurations use ASR transcripts. For reference, the same systems with ground-truth (GT) transcripts in place of ASR output (Sys.3--6) are reported in Appendix~\ref{sec:iemocap_gt} (Table~\ref{tab:iemocap_gt}). For each input configuration we evaluate two variants of our speechLLM: one with generative decoding and one with discriminative readout through the CLS head (+CLS), as described in Section~\ref{section:methodology}. The CLS head yields a statistically significant improvement on the speech-only system (Sys.1\,$\rightarrow$\,Sys.2), showing that discriminative readout is advantageous even when the model has no access to transcripts. Adding the transcript gives the largest gain (Sys.2\,$\rightarrow$\,Sys.4, also significant), and all multimodal configurations (Sys.3--6) surpass both speech-only variants and every unimodal baseline in Table~\ref{tab:unimodal_baselines} (except Audio Flamingo 3), confirming that acoustic and linguistic modalities are complementary. Conversational context adds a slight further gain, reaching 76.97 Macro F1. Our best UA (77.50) matches the strongest comparable result in the literature (77.60, \citealp{chen25i_interspeech}), which requires a far heavier pipeline --- extra disentanglement modules, a larger LLM (8B), two-stage multitask training, and 739\,h of external data --- whereas ours is a single linear head on a frozen 3B backbone with no external data. The large standard deviations reflect speaker difficulty, as each fold holds out a distinct speaker pair. Per-class F1 gains from each generative system to its discriminative counterpart are positive throughout (Angry +1.40, Happy +2.37, Neutral +2.13, Sad +2.29 on average), confirming that the gain is class-balanced.

% Table~\ref{tab:asr_results} reports results in the realistic ASR transcript setting, where the CLS head yields the most substantial gains (Macro F1 improves from 74.47 to 76.50 for the single encoder), suggesting that discriminative supervision is especially beneficial when input quality is degraded. Furthermore, by bypassing the token generation process entirely, the CLS head eliminates the risk of hallucinated emotion label predictions. Here transcripts are generated using a fine-tuned Whisper-medium model (WER: 8.06\%). The CLS head variant demonstrates robustness to noisy text inputs. This is practically significant, as most prior MER studies assume oracle transcripts, a condition rarely met in deployment.

Table~\ref{tab:main_results_meld} reports our MELD results. As on IEMOCAP, the CLS head gives a statistically significant improvement on the speech-only system (Sys.1\,$\rightarrow$\,Sys.2). With ASR transcripts, however, it only matches generative decoding rather than improving on it (Sys.3--6), and all four configurations fall considerably short of prior multimodal baselines, which use GT transcripts. We therefore repeated the multimodal configurations with GT transcripts, both for comparability with prior work and to isolate the effect of ASR: Sys.7--10 mirror Sys.3--6 in order, with \textit{text} and \textit{context} taken from GT transcripts rather than ASR output. Under GT transcripts the IEMOCAP trends reappear: the CLS head is the better variant in every configuration, and the speech-to-multimodal gain is large and significant (Sys.2\,$\rightarrow$\,Sys.8, 33.32\,$\rightarrow$\,48.59 Macro F1), with per-class F1 gains from Sys.7 to Sys.8 spread evenly across classes and largest on the minority emotions (Disgust +5.58, Fear +4.62, Sadness +2.96). We attribute the cost of substituting ASR output for GT transcripts (best WA 66.04 against 54.81) to transcript quality. MELD utterances were cut from \textit{Friends} episodes by an auto-aligner driven by subtitle timestamps, so a considerable number of segments are misaligned with their reference text,\footnote{\url{https://github.com/declare-lab/MELD/issues/9}} which drives our ASR to 32.38\% WER. Inspecting the data, we found many clips cut so short that they yield an empty hypothesis despite a far longer reference transcript, so removing GT transcripts strips away information the model would otherwise have. The unimodal baselines on MELD indicate the same: text baseline with access to GT transcripts surpass strongest speech baseline by a wide margin (47.32 against 43.33 Macro F1; Table~\ref{tab:unimodal_baselines}). To confirm that the degradation stems from ASR quality, we repeated Sys.6 on the subset of utterances with WER\,$<$\,100\%: it rises from 54.53 to 57.48 WA and overtakes its generative counterpart (Sys.5, 55.53 WA on the same subset). Given how drastically the two conditions differ, the gains claimed by prior MELD baselines, which report considerably better metrics using GT transcripts, might not hold up in a realistic ASR setting; hence the importance of evaluating on ASR transcripts, as we do. Even so, our best multimodal system remains ahead of the strongest unimodal baseline (48.77 against 47.32 Macro F1), confirming that acoustic and linguistic cues stay complementary.

\subsection{ Ablation: Frozen LLM (no LoRA)}
\label{subsection:ablation_training}
Table~\ref{tab:ablation_base} ablates the training configuration. As described in Section~\ref{sec:speechllm_implementation}, our main setup trains the speech projector, the CLS head and the LoRA components of the LLM. Here LoRA is disabled, so the generative variant trains only the speech projector, and the discriminative variant (+ CLS) the projector together with the CLS head. The CLS head improves every configuration except the challenging MELD ASR setting, by up to $+3.77$ WA on IEMOCAP: a negligible number of extra trainable parameters on top of the projector yields a sizeable gain, making discriminative readout attractive whenever the LLM cannot be fine-tuned or compute is constrained. We further observed a slight rise in hallucinations once the LLM is left frozen, which the CLS head eliminates by construction.

%ABLATION: BASE-ONLY TRAINING
\begin{table*}[t!]
% continuous vertical rules: remove booktabs' gaps above/below horizontal rules,
% and restore the lost breathing room with extra row height
\setlength{\aboverulesep}{0pt}
\setlength{\belowrulesep}{0pt}
\setlength{\extrarowheight}{.4ex}
\setlength{\tabcolsep}{3pt}
\renewcommand{\arraystretch}{1.0}
\centering
\footnotesize
\caption{Ablation: Frozen LLM (No LoRA). All configurations take \textit{audio} as input and use both the current utterance's transcript and a single preceding utterance as context. \textbf{ASR} characterises the textual input: \cmark{} for ASR transcripts and \xmark{} for ground-truth transcripts; \textbf{Context} is transcribed in the same way. Results are mean\,$\pm$\,std over the 5 folds (IEMOCAP) and 3 random seeds (MELD); \textbf{bold} marks the best result in each configuration, per dataset.}
\label{tab:ablation_base}
\begin{tabular}{l|l|c|c|c|c|c|c|c|c}
\toprule
\multirow{2}{*}{\textbf{ID}} & \multirow{2}{*}{\textbf{Method}}
& \multicolumn{2}{c|}{\textbf{Textual Inputs}}
& \multicolumn{3}{c|}{\textbf{IEMOCAP}}
& \multicolumn{3}{c}{\textbf{MELD}} \\
\cmidrule(r){3-4} \cmidrule(lr){5-7} \cmidrule(l){8-10}
& & \textbf{ASR} & \textbf{Context}
& \textbf{WA} & \textbf{UA} & \textbf{Macro F1}
& \textbf{WA} & \textbf{W-F1} & \textbf{Macro F1} \\
\midrule
1 & Ours
& \xmark & \cmark
& 71.46\,{\scriptsize$\pm$4.60} & 72.34\,{\scriptsize$\pm$3.85} & 72.06\,{\scriptsize$\pm$4.88}
& 60.50\,{\scriptsize$\pm$0.48} & 59.39\,{\scriptsize$\pm$0.43} & 42.64\,{\scriptsize$\pm$1.16} \\
2 & Ours (+ CLS)
& \xmark & \cmark
& \textbf{75.23}\,{\scriptsize$\pm$2.80} & \textbf{76.01}\,{\scriptsize$\pm$3.03} & \textbf{75.55}\,{\scriptsize$\pm$2.67}
& \textbf{62.11}\,{\scriptsize$\pm$0.82} & \textbf{60.55}\,{\scriptsize$\pm$0.65} & \textbf{43.47}\,{\scriptsize$\pm$0.93} \\
\cmidrule{1-10}
3 & Ours
& \cmark & \cmark
& 72.20\,{\scriptsize$\pm$2.36} & 72.64\,{\scriptsize$\pm$2.38} & 72.62\,{\scriptsize$\pm$2.58}
& \textbf{54.06}\,{\scriptsize$\pm$0.47} & \textbf{52.67}\,{\scriptsize$\pm$0.54} & \textbf{36.36}\,{\scriptsize$\pm$0.73} \\
4 & Ours (+ CLS)
& \cmark & \cmark
& \textbf{74.16}\,{\scriptsize$\pm$3.23} & \textbf{74.96}\,{\scriptsize$\pm$3.54} & \textbf{74.63}\,{\scriptsize$\pm$3.27}
& 53.35\,{\scriptsize$\pm$0.84} & 51.72\,{\scriptsize$\pm$1.18} & 34.58\,{\scriptsize$\pm$1.63} \\
\bottomrule
\end{tabular}
\end{table*}

% \section{Interpretability: What the LLM Encodes}
% \label{subsection:interpretability}

\section{Vocabulary-Space Analysis of the Class Vectors}
\label{sec:interpretability}

\begin{figure}[t]
  \centering
  \includegraphics[width=\linewidth]{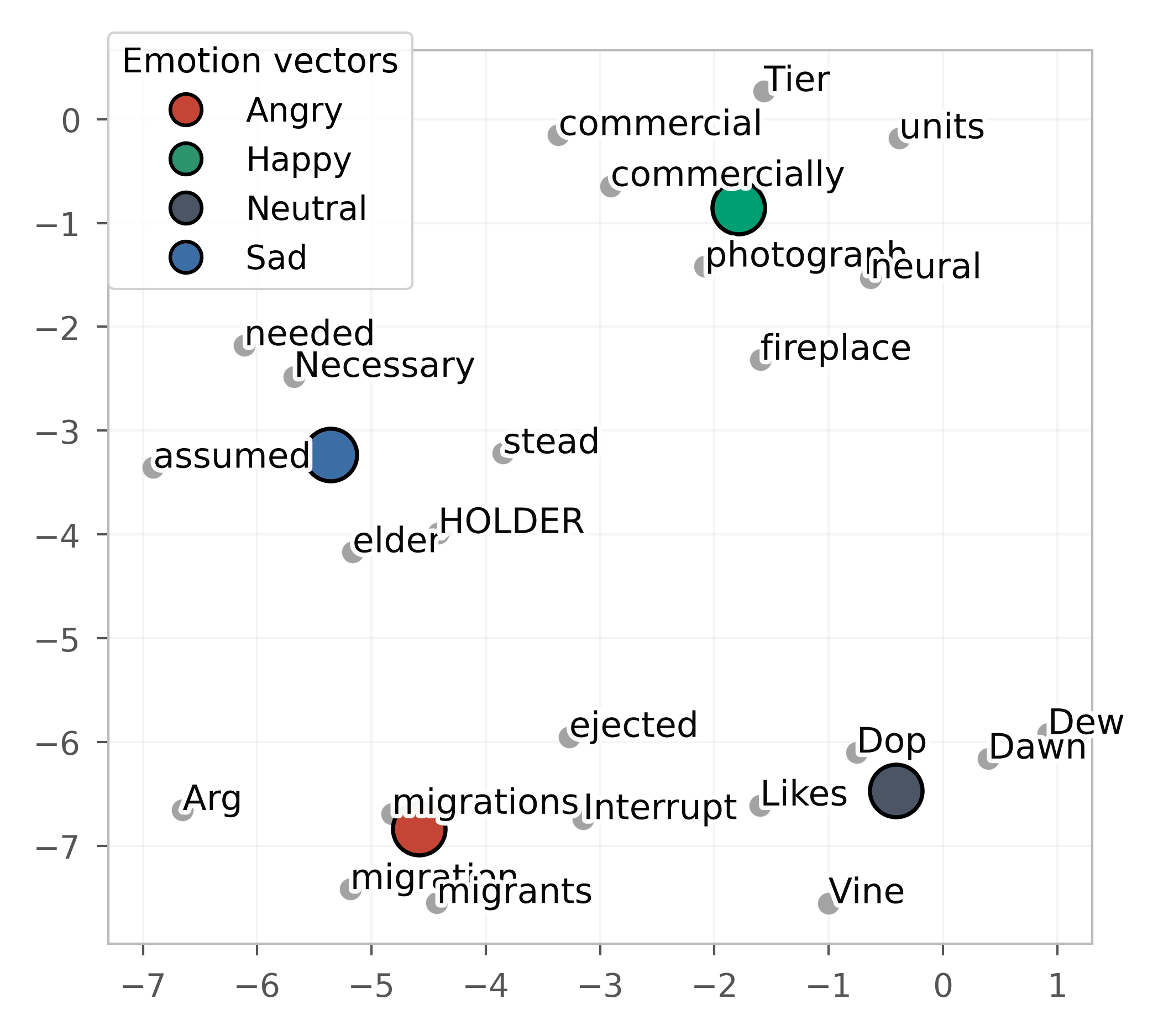}
  \caption{Two-dimensional t-SNE projection of the LLM token-embedding space, for a single IEMOCAP checkpoint. Each colored circle marks one learned emotion class vector (a column of $\mathbf{W}_{\text{CLS}}$); surrounding grey labels are token embeddings, with the top-$k$ tokens recovered for each emotion shown near its vector. Every emotion vector sits at the center of its own token cloud (not literal affect words, but indirect associations), illustrating that the linear head groups tokens by emotion.}
  \label{fig:tsne}
\end{figure}

As described in Section~\ref{section:CLS_head}, because the CLS head is linear, each emotion's class vector lives in the same space as the LLM's output token embeddings and is directly comparable to them. We use this to ask \emph{what} it has learned each emotion to be --- which tokens it has associated with each emotion --- internally, in the token space. For every trained checkpoint, across all input configurations and all folds (IEMOCAP) and seeds (MELD), we compute the dot-product similarity between each class vector and every token embedding and read off the top-$k$ most similar tokens. The trends reported below recur across all of them; top-15 token lists for multiple input configurations, folds and seeds on both corpora are given in Appendix~\ref{sec:token_lists}. These similarities are associations in the output space: they do not show that the listed tokens drive the model's predictions. What they do show is systematic structure rather than noise --- the same lexical neighbourhoods recur across input configurations and folds, pointing to regularities the LLM carries from its pretraining rather than to artifacts. Figure~\ref{fig:tsne} makes this concrete: projected to two dimensions with t-SNE, each emotion vector is surrounded by its own cloud of associated tokens, so the head anchors a distinct lexical neighborhood per emotion.

On IEMOCAP, the recovered tokens are almost never literal affect words (\emph{angry}, \emph{sad}, \emph{happy}); instead each emotion resolves to a coherent but indirect semantic field. \textbf{Sad} is dominated by necessity- and deficit-related language (\emph{needed}, \emph{necessary}, \emph{assumed}) together with ageing cues (\emph{aging}, \emph{elder}), consistent with a field centred on lack and reduced agency. \textbf{Angry} aligns most strongly with dispute and confrontation vocabulary (\emph{argument}, \emph{arg}, \emph{conflict}, \emph{interrupt}, \emph{disruption}, \emph{dismissal}, \emph{withdraw}), alongside \emph{migration} and a water-related group (\emph{sea}, \emph{swim}, \emph{waters}) for which we see no evident affective reading. \textbf{Happy} maps onto reward and elevated-positive language (\emph{dopamine}, \emph{vine}, \emph{Kirk}, \emph{epic}, \emph{divine}), with \emph{dawn} and \emph{dew} adding positive imagery. \textbf{Neutral}, by contrast, clusters around transactional and categorial vocabulary (\emph{profession}, \emph{commercially}, \emph{tier}, \emph{units}) and plain object or place descriptors (\emph{photograph}, \emph{playground}), consistent with the near-absence of emotional loading.

On MELD: for several emotions the top tokens are the literal emotion words. \textbf{Fear} is almost entirely explicit (\emph{nervous}, \emph{terrified}, \emph{frightened}, \emph{anxiety}, \emph{panicked}, \emph{fragile}), as are \textbf{surprise} (\emph{surprise}, \emph{surprised}, \emph{surprising}, \emph{astonished}, \emph{wonder}) and \textbf{disgust} (\emph{disgust}, \emph{disgusting}, \emph{ugly}, \emph{dirt}). The remaining classes resolve to indirect fields: \textbf{anger} to dispute vocabulary (\emph{argument}, \emph{arg}, \emph{conflicts}, \emph{wrath}, \emph{interrupt}), \textbf{joy} to blessing and affection (\emph{bless}, \emph{loved}, \emph{Fiesta}), \textbf{sadness} to heaviness and grief (\emph{heavy}, \emph{sob}, \emph{lament}/\emph{mourning}), and \textbf{neutral} to flat everyday vocabulary with no emotional loading (\emph{unavoid}, \emph{viability}, \emph{hometown}, \emph{starters}, \emph{traveler}).

Overall, these patterns hold across input configurations, folds/seeds and partly recur across corpora (dispute vocabulary for anger), so they are a property of the learned class vectors rather than noise: the model grounds emotion recognition less in explicit affect words than in broader semantic priors acquired during pretraining, including associations in sensitive domains. The linear CLS head is what surfaces them, tying each emotion to a single direction in the token space so that these associations can be read directly off the class vectors rather than inferred from downstream behavior.

\section{Conclusions}

We showed that a linear classification head is enough to adapt a speechLLM for emotion recognition, and how this lightweight adaptation can be used to navigate the conditions a deployed system encounters, such as ground-truth versus ASR transcripts. Being complementary to any LLM-based system, it should carry over to larger speechLLMs, which may yield stronger results. Across two benchmarks it gives significant gains where ASR quality is sufficient, and is likewise beneficial under constrained compute. Its linearity lets us examine the correlations it has learned in the LLM's output token space.

\section*{Limitations}

\begin{itemize}
\item \textbf{Single architecture and backbone.} All experiments use one speechLLM: a frozen Whisper-medium encoder, a two-layer projector and LLaMA-3.2-3B with LoRA. We do not vary the speech encoder, the connector design or the size and family of the LLM. Because the discriminative readout acts on the final hidden state, its behaviour may differ for backbones whose output space is shaped differently by pretraining, and our absolute numbers are not comparable to systems built on larger LLMs.

\item \textbf{Limited coverage of ASR conditions.} We characterise transcript quality with a single ASR system at two operating points (8.06\% WER on IEMOCAP, 32.38\% on MELD). Our conclusion that the classification head helps when ASR quality is sufficient therefore rests on two measurements, and we cannot say where the threshold lies, how it depends on the error type rather than the error rate, or whether a stronger ASR system would recover the gains lost on MELD.

\item \textbf{Corpora and language.} We evaluate on IEMOCAP and MELD, both English and neither spontaneous: IEMOCAP is acted dyadic interaction and MELD is scripted television dialogue, with known audio--transcript misalignment. Our findings may not transfer to spontaneous speech, to other languages, or to label sets other than the 4-class and 7-class formulations used here. We also provide context as a single preceding utterance, so we do not test longer dialogue history or speaker-aware context.

% \item \textbf{Statistical power.} Significance is assessed over 5 cross-validation folds on IEMOCAP and 3 random seeds on MELD. These are small samples for a paired test, and the MELD comparisons in particular should be read as indicative rather than conclusive.

\item \textbf{Scope of the vocabulary-space analysis.} The token lists are correlations between the learned class vectors and the LLM's output embeddings. They do not establish that the listed tokens influence predictions, and we perform no intervention to test this. We report the top-15 tokens under dot-product similarity; other ranking criteria may surface different neighbourhoods. Several recurring groups, such as the water-related tokens for anger, have no interpretation we can offer, and we report them rather than omit them.
\end{itemize}

% \section*{Ethical Considerations}

\section*{Acknowledgments}

This work was supported by Idiap Research Institute and Uniphore collaboration project. It was also partially funded by the SNSF through the SPIRIT project \textbf{ORIENTER}: tOwards undeRstanding and modelIng the language of mENTal health disordERs (grant no. IZSTZ0\_223488).

% \bibliography{custom,anthology-overleaf-1,anthology-overleaf-2}

\bibliography{custom}

\appendix

\section{Unimodal Baselines}
\label{sec:unimodal_baselines}

Table~\ref{tab:unimodal_baselines} reports the full unimodal baseline results summarised in Section~\ref{section:results}, covering the Speech Foundation Models, the zero-shot text-only LLM and the off-the-shelf multimodal LLMs described in Section~\ref{section:SER_baselines}, on both corpora.

%UNIMODAL BASELINES TABLE
\begin{table*}[t!]
% continuous vertical rules: remove booktabs' gaps above/below horizontal rules,
% and restore the lost breathing room with extra row height
\setlength{\aboverulesep}{0pt}
\setlength{\belowrulesep}{0pt}
\setlength{\extrarowheight}{.4ex}
\setlength{\tabcolsep}{5pt}
\renewcommand{\arraystretch}{1.0}
\centering
\footnotesize
\caption{Unimodal baselines with Speech Foundation Models (SFM), LLMs and multimodal LLMs (MLLM) on IEMOCAP and MELD. All models take \textit{audio} as input except those marked ``$\circ$'', which are text-only. \textbf{ASR} characterises the textual input: \cmark{} for ASR transcripts, \xmark{} for ground-truth transcripts, and ``--'' when the model uses no text data at all. \textbf{Context} indicates whether preceding textual context is provided. Best result in each block is highlighted in \textbf{bold}. $^{\dagger}$Best entry on the standard EmoBox leaderboard, shown for reference and directly comparable to our Whisper-large result below.}
\label{tab:unimodal_baselines}
\begin{tabular}{l|c|c|c|c|c|c|c|c}
\toprule
\multirow{2}{*}{\textbf{Model}}
& \multicolumn{2}{c|}{\textbf{Textual Inputs}}
& \multicolumn{3}{c|}{\textbf{IEMOCAP}}
& \multicolumn{3}{c}{\textbf{MELD}} \\
\cmidrule(r){2-3} \cmidrule(lr){4-6} \cmidrule(l){7-9}
& \textbf{ASR} & \textbf{Context}
& \textbf{WA} & \textbf{UA} & \textbf{Macro F1}
& \textbf{WA} & \textbf{W-F1} & \textbf{Macro F1} \\
\midrule

% ---- SFM ----
\rowcolor{gray!15}
\multicolumn{9}{l}{\textit{SFM}} \\
\midrule
Whisper-large$^{\dagger}$ & -- & -- & \textbf{72.86} & \textbf{73.54} & \textbf{73.11} & 51.89 & -- & \textbf{32.95} \\
Whisper-large             & -- & -- & 70.28 & 70.63 & 70.45 & \textbf{55.13} & \textbf{49.00} & 27.80 \\
WavLM-large               & -- & -- & 68.33 & 68.24 & 68.38 & 52.34 & 45.09 & 24.29 \\
HuBERT-base               & -- & -- & 64.98 & 67.16 & 66.03 & 50.38 & 43.01 & 21.50 \\
\midrule

% ---- LLM ----
\rowcolor{gray!15}
\multicolumn{9}{l}{\textit{LLM}} \\
\midrule
Gemma3-27b$^{\circ}$ & \xmark & -- & \textbf{55.39} & \textbf{53.81} & \textbf{54.16} & \textbf{57.28} & \textbf{58.77} & \textbf{47.32} \\
\midrule

% ---- MLLM ----
\rowcolor{gray!15}
\multicolumn{9}{l}{\textit{MLLM}} \\
\midrule
Audio Flamingo 3 & -- & -- & \textbf{76.48} & \textbf{76.15} & \textbf{76.69} & \textbf{58.58} & \textbf{55.74} & 40.01 \\
Qwen2.5-Omni-7B  & -- & -- & 70.46 & 70.38 & 70.98 & 53.30 & 51.25 & 35.25 \\
Qwen3-Omni-30B   & -- & -- & 67.33 & 66.46 & 67.54 & 53.56 & 54.02 & \textbf{43.33} \\
MiniCPM-o-4.5    & -- & -- & 48.47 & 43.96 & 43.48 & 54.95 & 49.07 & 32.10 \\
Phi4-Multimodal  & -- & -- & 47.30 & 48.44 & 46.83 & 31.95 & 30.38 & 25.52 \\
\bottomrule
\end{tabular}
\end{table*}

\section{IEMOCAP Results with Ground-Truth Transcripts}
\label{sec:iemocap_gt}

Table~\ref{tab:iemocap_gt} repeats the IEMOCAP configurations of Table~\ref{tab:main_results_iemocap} with ground-truth transcripts in place of ASR output: Sys.3--6 use human transcripts for both the current utterance and its context, while Sys.1--2 use no text and are therefore identical to the main table.

%IEMOCAP GROUND-TRUTH-TRANSCRIPT RESULTS (APPENDIX)
\begin{table*}[t!]
% continuous vertical rules: remove booktabs' gaps above/below horizontal rules,
% and restore the lost breathing room with extra row height
\setlength{\aboverulesep}{0pt}
\setlength{\belowrulesep}{0pt}
\setlength{\extrarowheight}{.4ex}
\setlength{\tabcolsep}{6pt}
\renewcommand{\arraystretch}{1.0}
\centering
\footnotesize
\caption{Results on IEMOCAP with ground-truth transcripts. All configurations take \textit{audio} as input. \textbf{ASR} characterises the textual input: \xmark{} for ground-truth transcripts, and ``--'' when no text data is used. \textbf{Context} indicates whether preceding textual context is provided. Results are mean\,$\pm$\,std over the 5 folds; \textbf{bold} marks the best result in each configuration. ``$*$'' and ``$\dagger$'' represent statistically significant (Paired Single-tailed $t$-test, $p=0.05$) improvements over Sys.1 and Sys.2, respectively.}
\label{tab:iemocap_gt}
\begin{tabular}{l|l|c|c|c|c|c}
\toprule
\multirow{2}{*}{\textbf{ID}} & \multirow{2}{*}{\textbf{Method}}
& \multicolumn{2}{c|}{\textbf{Textual Inputs}}
& \multicolumn{3}{c}{\textbf{IEMOCAP}} \\
\cmidrule(r){3-4} \cmidrule(l){5-7}
& & \textbf{ASR} & \textbf{Context}
& \textbf{WA} & \textbf{UA} & \textbf{Macro F1} \\
\midrule
1 & Ours
& -- & --
& 69.86\,{\scriptsize$\pm$2.36} & 70.50\,{\scriptsize$\pm$2.43} & 70.06\,{\scriptsize$\pm$2.63} \\
2 & Ours (+ CLS)
& -- & --
& \textbf{72.35}\,{\scriptsize$\pm$1.80} & \textbf{73.14}\,{\scriptsize$\pm$2.14} & \textbf{72.80}$^{*}$\,{\scriptsize$\pm$2.06} \\
\cmidrule{1-7}
3 & Ours
& \xmark & --
& 76.06\,{\scriptsize$\pm$2.01} & 76.65\,{\scriptsize$\pm$1.96} & 76.33\,{\scriptsize$\pm$1.63} \\
4 & Ours (+ CLS)
& \xmark & --
& \textbf{76.40}\,{\scriptsize$\pm$3.37} & \textbf{77.21}\,{\scriptsize$\pm$3.74} & \textbf{76.97}$^{\dagger}$\,{\scriptsize$\pm$3.29} \\
\cmidrule{1-7}
5 & Ours
& \xmark & \cmark
& \textbf{78.34}\,{\scriptsize$\pm$3.55} & \textbf{79.83}\,{\scriptsize$\pm$2.67} & \textbf{78.80}\,{\scriptsize$\pm$3.45} \\
6 & Ours (+ CLS)
& \xmark & \cmark
& 77.36\,{\scriptsize$\pm$4.04} & 77.68\,{\scriptsize$\pm$5.36} & 77.51$^{\dagger}$\,{\scriptsize$\pm$4.78} \\
\bottomrule
\end{tabular}
\end{table*}

\section{Vocabulary-Space Analysis of the Class Vectors}
\label{sec:token_lists}

Table~\ref{tab:token_lists} lists the top-15 tokens for each emotion class vector on IEMOCAP, for four checkpoints: one trained with speech and transcript, and three folds of the configuration that additionally uses conversational context. The lists are stable both across folds of a given configuration and across configurations, and the same groups recur in morphological variants (\emph{Arg}/\emph{args}/\emph{Argument}, \emph{migration}/\emph{migrants}/\emph{migrating}, \emph{Sea}/\emph{waters}/\emph{aquatic}/\emph{Swim}). The assignment is also unambiguous: in all five folds, every one of the 15 tokens listed for an emotion scores highest against that emotion and against no other, so the lists are not an artefact of mixed-up class indices.

%TOP-15 TOKEN LISTS (APPENDIX)
\begin{table*}[t!]
\setlength{\aboverulesep}{0pt}
\setlength{\belowrulesep}{0pt}
\setlength{\extrarowheight}{.4ex}
\setlength{\tabcolsep}{3pt}
\renewcommand{\arraystretch}{1.0}
\centering
\footnotesize
\caption{Top-15 tokens per emotion on IEMOCAP, ranked by dot-product similarity between each class vector and the LLM output token embeddings. Rows cover one checkpoint trained with speech and transcript (F1) and three folds of the configuration that additionally uses conversational context (F2, F4, F5).}
\label{tab:token_lists}
\begin{tabular}{p{52pt}|p{89pt}|p{89pt}|p{89pt}|p{89pt}}
\toprule
\textbf{Setting} & \textbf{Angry} & \textbf{Happy} & \textbf{Neutral} & \textbf{Sad} \\
\midrule
Speech + text (F1)
& Arg, dismissal, migration, Interr, waters, camb, footh, Bobby, Sea, migrants, migrating, disruption, seas, araya, Interrupt
& Dew, Kirk, Dawn, alle, Vine, Lin, fen, bou, Shadows, UNESCO, orange, Dop, epic, singly, MLP
& Tier, commercially, commercial, photograph, units, heater, airborne, Verb, Profession, caz, Nation, periods, cerebral, automobile, Playground
& Elder, Sext, stead, needed, Necessary, overt, aging, assumed, dressing, Kraj, Eag, Dixon, Estr, Eig, HOLDER \\
\cmidrule{1-5}
Speech + text + context (F2)
& Arg, migration, Argument, Interrupt, dismissal, relocation, args, Vac, footh, Interr, routes, yourself, tuo, Fallon, unmarried
& fen, Dew, Dod, Kirk, dop, Bou, Milo, Vine, dopamine, Lin, Divine, Shadows, epic, Swan, bunch
& Tier, photograph, Gee, Profession, units, neural, nervous, Verb, Railroad, Cost, caz, commercial, Nation, commercially, Tale
& needed, assumed, Elder, Sext, Kraj, Necessary, stead, Kop, dressing, Germans, overt, Cars, HOLDER, required, Dress \\
\cmidrule{1-5}
Speech + text + context (F4)
& Arg, waters, migration, press, crossover, literature, aquatic, water, upright, sea, Swim, args, unmarried, Single, kon
& Dew, Fran, Dod, Vel, Lin, Vine, Fried, Arts, Kirk, Dawn, Sketch, Rep, Shadows, streak, fen
& Profession, photograph, Tier, Quad, commercially, caz, units, Playground, Saturdays, roz, predatory, tightened, sez, heater, Verb
& assumed, perceived, stead, Kraj, aging, Elder, dressing, Dress, Sanat, Sez, fran, Latter, overt, shadow, Sext \\
\cmidrule{1-5}
Speech + text + context (F5)
& Arg, migration, args, footh, Sea, withdraw, Argument, crossover, araya, connectors, outdoors, camb, dismissal, migrating, migr
& Dew, fen, Kirk, Sketch, UNESCO, epic, Lin, Vine, Shadows, Dawn, bunch, bou, whipped, orange, Shapiro
& Tier, units, caz, commercial, Gee, commercially, Profession, Verb, Cost, photograph, stove, Tale, Nation, cent, Point
& needed, assumed, stead, Elder, Kraj, aging, perceived, Necessary, Estr, Sext, Germans, Dress, expected, overt, required \\
\bottomrule
\end{tabular}
\end{table*}

Table~\ref{tab:token_lists_meld} gives the corresponding lists for MELD, for the two input configurations trained on that corpus. The lists are stable across the two configurations and recur in morphological variants (\emph{Arg}/\emph{Args}/\emph{Argument}, \emph{disgust}/\emph{disgusting}/\emph{disgusted}, \emph{surprise}/\emph{surprises}/\emph{surprised}/\emph{surprising}). Unlike IEMOCAP, the vectors for fear, surprise and disgust are headed by the literal emotion words themselves.

%TOP-15 TOKEN LISTS, MELD (APPENDIX)
\begin{table*}[t!]
\setlength{\aboverulesep}{0pt}
\setlength{\belowrulesep}{0pt}
\setlength{\extrarowheight}{.4ex}
\setlength{\tabcolsep}{3pt}
\renewcommand{\arraystretch}{1.0}
\centering
\footnotesize
\caption{Top-15 tokens per emotion on MELD, ranked by dot-product similarity between each class vector and the LLM output token embeddings, for the two input configurations trained on MELD.}
\label{tab:token_lists_meld}
\begin{tabular}{l|p{192pt}|p{192pt}}
\toprule
\textbf{Emotion} & \textbf{Speech + text} & \textbf{Speech + text + context} \\
\midrule
Anger & Sea, Interrupt, Single, Sense, camb, erg, unmarried, Argument, Arg, Swim, aggress, withdraw, conflicts, Sec, wrath & Arg, Argument, kon, Sea, Args, araya, argparse, wrath, walker, prec, Fri, conflicts, Interrupt, migration, Sean \\
\cmidrule{1-3}
Disgust & disgust, ugl, Kirk, Dew, deg, disgusted, UNESCO, dislike, disgusting, Dirt, Dre, grotes, sketch, ugly, textures & disgust, ugl, Dew, Kirk, deg, disgusting, Lar, staining, filtro, Dirt, Fran, ugly, UNESCO, coating, Bod \\
\cmidrule{1-3}
Fear & nervous, terrified, nerv, anxiety, anxious, frightened, panicked, fearing, stolen, crippling, paranoid, caz, fragile, Dread, Quad & nervous, frightened, terrified, nerv, anxious, anxiety, fearing, panicked, nerves, afraid, fearful, commercially, fragile, scared, tightening \\
\cmidrule{1-3}
Joy & overt, bubb, Dress, dressing, Fiesta, Elder, Kraj, assumed, Sext, loved, excitement, Bless, Eag, Commonwealth, Sez & overt, dressing, Kraj, bubb, Dress, Bless, Fiesta, Recommend, blessing, celestial, loved, wonderful, love, Loving, celebration \\
\cmidrule{1-3}
Neutral & alm, advis, savory, viability, unavoid, afirm, hometown, subs, Hlav, starters, tomorrow, revival, lifes, traveler, Ebony & starters, traveler, unavoid, latex, Rubber, Lyft, Ebony, alm, newcom, viability, hometown, shoe, inevitably, realism, RVA \\
\cmidrule{1-3}
Sadness & Physiology, suppl, Jug, wall, trib, heavy, Gund, Bauer, sob, Barcl, pau, lament, beden, Carpet, BITS & Gund, jug, Physiology, heavy, suppl, sob, mourning, SOCKET, pau, bus, wall, StatusBar, Sidebar, trib, CAMERA \\
\cmidrule{1-3}
Surprise & surprises, wonder, surprise, aston, incred, stun, vap, bewild, surprising, grades, astonished, odd, surprised, Watts, spo & surprises, surprising, unexpectedly, incred, surprised, aston, surprise, astonished, astonishing, wonder, surpr, exceptional, Watts, Oxygen, grades \\
\bottomrule
\end{tabular}
\end{table*}

\end{document}